\documentclass[10pt,twocolumn,letterpaper]{article}

\usepackage[pagenumbers]{wacv} 

\usepackage[T1]{fontenc}
\usepackage[utf8]{inputenc}
\usepackage{xcolor}
\usepackage[most]{tcolorbox}
\tcbuselibrary{breakable,skins}
\usepackage{caption}

\definecolor{promptborder}{RGB}{70,90,140}
\definecolor{promptbg}{RGB}{246,247,251}

\newtcolorbox{promptbox}[2]{
  enhanced, breakable,
  colback=promptbg, colframe=promptborder,
  boxrule=0.6pt, arc=1.5mm,
  fonttitle=\bfseries\small, coltitle=white, colbacktitle=promptborder,
  title={#1},
  left=4pt, right=4pt, top=4pt, bottom=4pt,
  label={#2},
}

\definecolor{wacvblue}{rgb}{0.21,0.49,0.74}
\usepackage[pagebackref,breaklinks,colorlinks,allcolors=wacvblue]{hyperref}

\def\wacvPaperID{*****} 
\def\confName{WACV}
\def\confYear{2027}

\title{Vision-language models see more about agriculture than they report, and test-time scaling remains bounded by parametric knowledge}

\author{
Earl Ranario \quad Jared Smith \quad Urmil Chandarana \quad Lars Lundqvist \quad Bharath Sridhar\quad Mason Earles \\[0.4em]
University of California, Davis \\
{\tt\small \{ewranario, jrssmith, uchandar, llund, bsridhar, jmearles\}@ucdavis.edu}
}

\begin{document}
\maketitle
\begin{abstract}
Vision-language models (VLMs) show promise for agricultural diagnosis but fail unpredictably in zero-shot settings. We test whether this stems from poor visual features or an inability to verbalize what the model already perceives, using a benchmark spanning classification and detection drawn from 118 AgML classification and 54 detection datasets. Linear probing shows agricultural features are linearly separable in frozen VLM vision towers, with Gemma-4-31B's encoder reaching F1-macro of 0.60--0.78 across tasks and approaching the vision-only DINOv3 backbone, indicating the bottleneck lies downstream of perception. We then test whether test-time scaling, an iterative in-context skill-memory loop, can close this gap without updating weights. Gains are modest and inconsistent. Skill-memory improves classification accuracy by up to 3.1 points on three of four tasks while regressing on species, with gains concentrated on classes seen during optimization and unseen classes regressing. Detection F1 is effectively unchanged (0.412 to 0.421), though an oracle filtering condition reaches 0.432 F1, revealing real headroom unconstrained prompting fails to surface. These results indicate VLMs see more agricultural features than they report, but test-time scaling alone fails to surface that hidden parametric knowledge, motivating agentic approaches grounded in external, retrievable knowledge.
\end{abstract}    
\section{Introduction}
\label{sec:intro}

Artificial intelligence (AI) drives innovation in modern agriculture, from automated machinery and robotic harvesting to precision farm management \cite{aijaz_artificial_2025}, including drone-based disease monitoring \cite{abbas_drones_2023} and crop breeding for genotype-environment evaluation \cite{berlingeri_integration_2025}. However, these methods remain limited by the availability of labeled data and resources \cite{jiang_convolutional_2020}. Efforts to expand publicly available agricultural datasets \cite{lu_survey_2020} may not adequately address multi-domain complexity. A model trained on one or more datasets does not guarantee generalization to another domain, given differences in lighting, camera angle, or plant species.

Foundation models (FMs) promise to be task-agnostic, pretrained on massive datasets to acquire general-purpose knowledge. Vision-language models (VLMs), a form of FM, are further trained to link visual representations to textual semantics \cite{radford_learning_2021}, but it remains unclear whether they are reliable enough for agricultural classification. Existing agricultural benchmarks reveal gaps between VLMs and domain-specific requirements \cite{joshi_standardizing_2023, shinoda_agrobench_2025}. As \cite{ranario_are_2026} notes, current VLMs are not yet agricultural-ready. While they show promise as assistive components, their zero-shot failures indicate a fundamental disconnect between perception and domain knowledge.

Closing this gap could proceed by further training the model on agricultural data, but no single adaptation strategy is uniformly favored across deployment circumstances. Fine-tuning requires labeled data and risks catastrophic forgetting or domain collapse, while test-time strategies avoid these costs by leaving model weights untouched but may recover a smaller or less reliable gain in return. This tradeoff, rather than an assumption that either approach is sufficient on its own, motivates the questions below.

\begin{figure}
  \centering
  \includegraphics[width=1.0\linewidth]{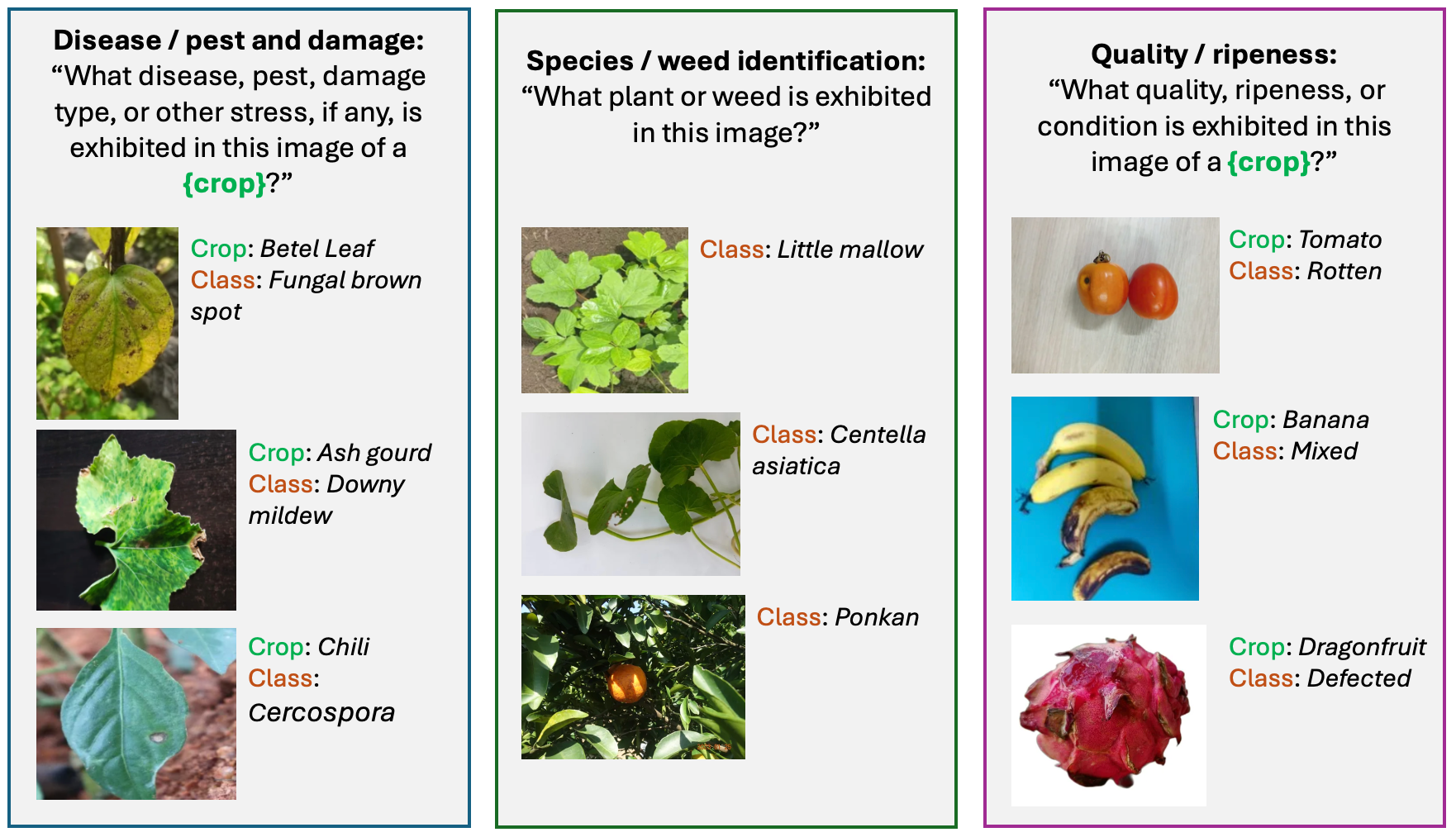}
  \caption{Classification questions used for each task type, shared across experiments.}
  \label{fig:baseline_prompts}
\end{figure}

\begin{figure*}[t]
\centering
\begin{subfigure}[b]{0.49\linewidth}
\includegraphics[width=\textwidth]{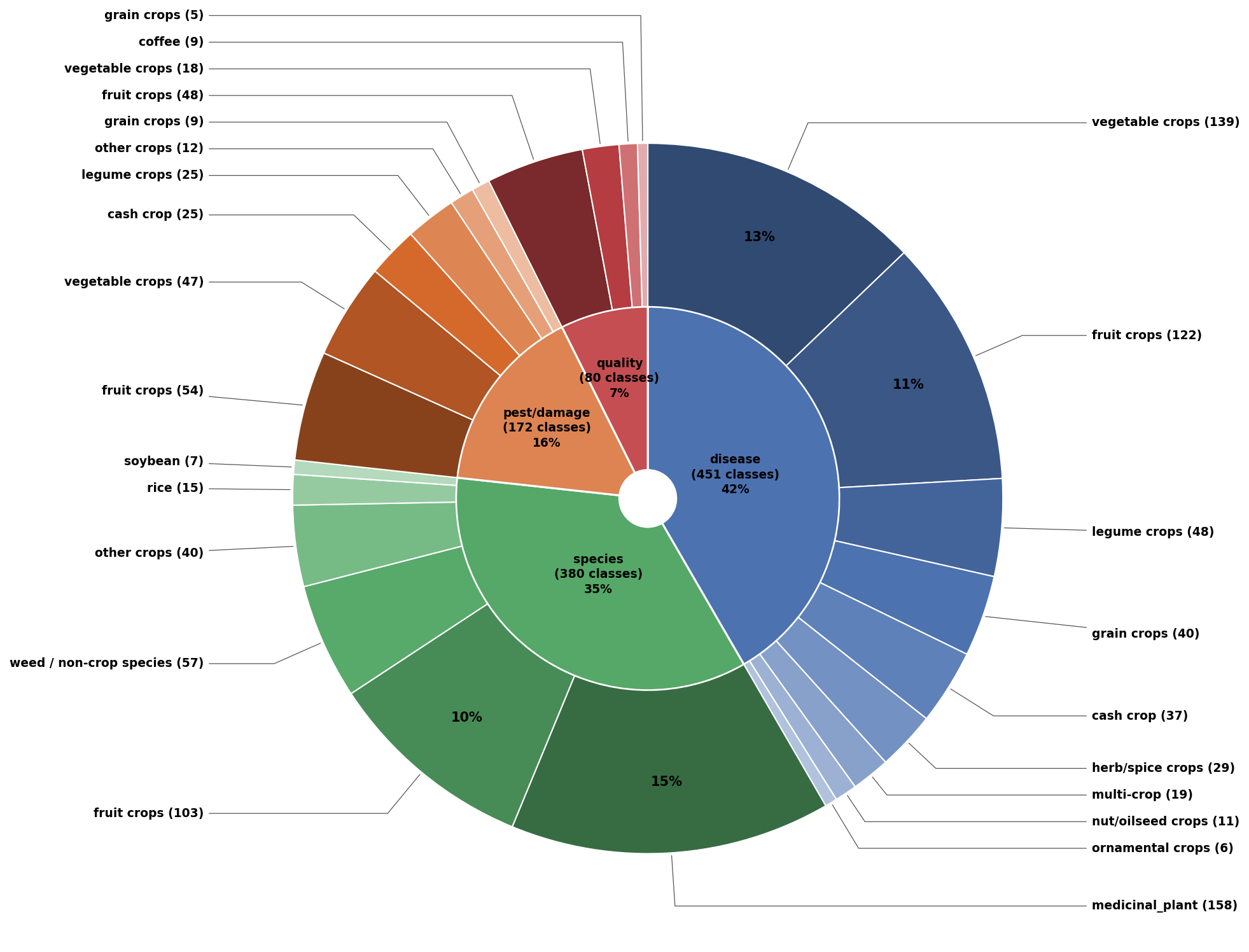}
\caption{By task}
\end{subfigure}
\hfill
\begin{subfigure}[b]{0.49\linewidth}
\includegraphics[width=\textwidth]{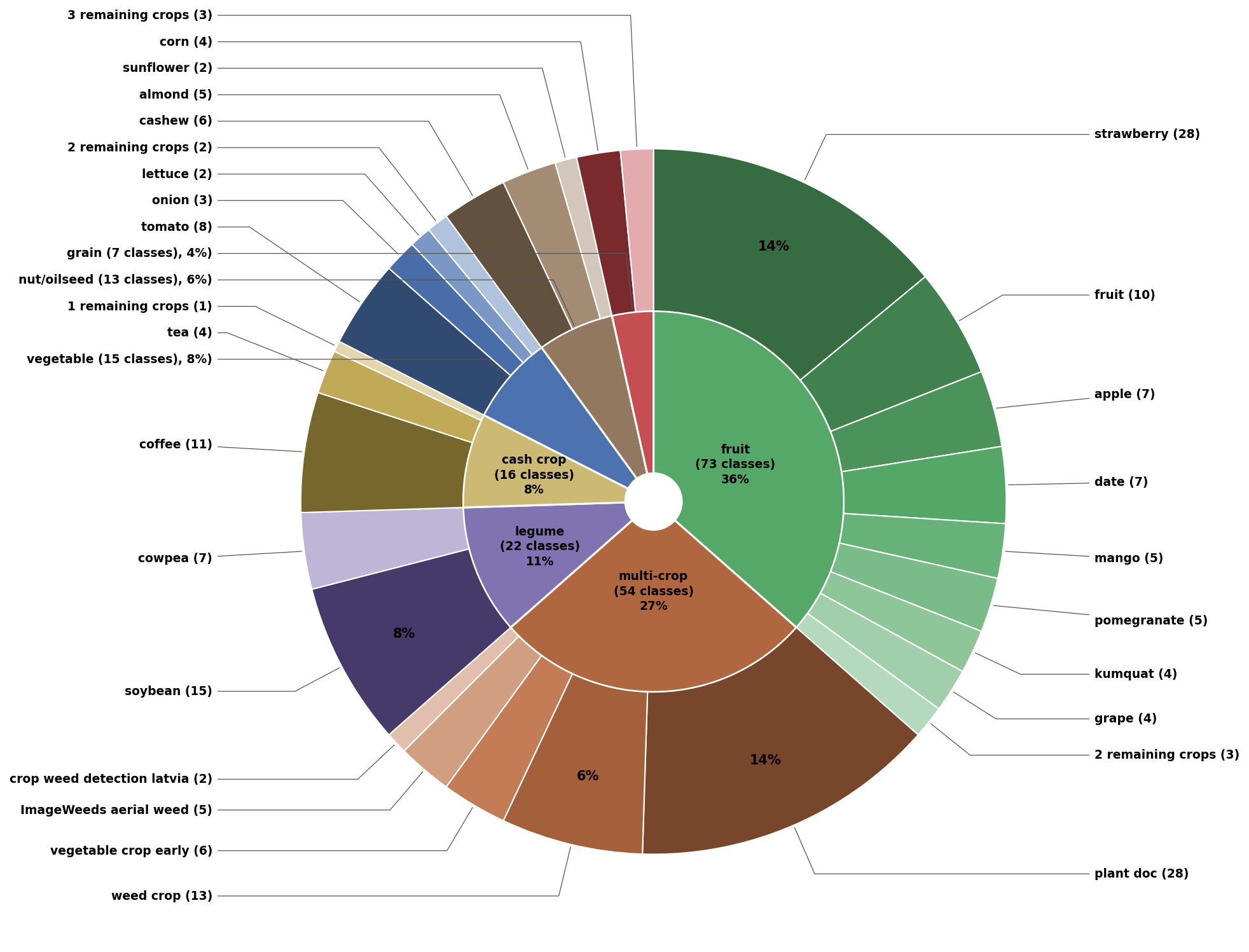}
\caption{By crop}
\end{subfigure}
\caption{Class distribution by crop category, drawn from the AgML collection: (a) 962
classes across 118 classification datasets, broken down by task type and crop; (b) 200
classes across 54 object detection datasets, broken down by crop.}
\label{fig:piecharts_nested}
\end{figure*}

\subsection{Research questions and hypotheses}
With that in mind, we ask: what drives the inability of current VLMs to zero-shot agricultural classification and detection tasks? Is it an inability to extract visual features from images, or to map those features to a diagnosis? Our hypotheses:

\begin{enumerate}
  \item \textit{Vision is good:} Vision towers encode sufficient fine-grained features for agricultural classification, but language components may lack the specific agricultural textual knowledge to generate reliable answers.
  \item \textit{Accumulated knowledge provides key context for predictions:} Accumulated knowledge can help identify complex features in images, motivated by how agricultural identification is typically taught and practiced: matching fine-grained physical characteristics against a written description, much like a field guide or dichotomous key.
\end{enumerate}

\subsection{Contributions} To answer these questions, we propose a diagnostic pipeline isolating and testing the vision and alignment components of VLMs on agricultural classification and detection benchmarks. Contributions include:

\begin{itemize}
  \item A benchmark suite constructed from the AgML data repository for image classification and object detection.
  \item Linear probing of VLM vision towers, showing agricultural features are linearly separable.
  \item Evaluation of test-time scaling techniques for agricultural tasks, from a rubric-guided chain-of-thought verifier to an iterative in-context skill-optimization framework.
\end{itemize}
\section{Datasets, models and background}
\label{sec:datasets}

Agricultural computer vision research has historically relied on individually released, narrowly scoped datasets, each covering a single crop, task, or region under its own imaging conditions, label taxonomy, and file format. This fragmentation makes systematic VLM evaluation difficult. A benchmark built from any one such dataset risks measuring a model's fit to that dataset's own idiosyncrasies (lighting, camera angle, regional disease naming) rather than its underlying diagnostic competence. AgML addresses this by aggregating a large number of independently published agricultural datasets into one standardized collection, making a single, broad-coverage benchmark across tasks and domains practical, rather than requiring separate, incompatible evaluations against dozens of individually sourced datasets.

All images were published into the AgML collection on the Hugging Face Hub (\url{https://huggingface.co/Project-AgML}) \cite{huggingface_website, joshi_standardizing_2023}, a curated aggregation of publicly released agricultural datasets spanning disease diagnosis, pest and physical/abiotic damage, and quality and species identification across crops and locations. We draw 118 classification datasets (962 classes, 13,350 images) and 54 object detection datasets (200 classes, 2,140 images) from this collection.\footnote{Download scripts and lessons/validation split manifests: \url{https://github.com/plantai-biophysics-lab/agtuned-vlm}}

We use DINOv3 (facebook/dinov3-vitb16-pretrain-lvd1689m) \cite{simeoni_dinov3_2025} and a YOLOv11 classification backbone (yolov11x-cls, Ultralytics) \cite{khanam_yolov11_2024} as frozen, purely vision-trained encoders for the linear-probe analysis. Alongside these, we probe four VLM vision towers spanning two model families and sizes: Qwen3.5-9B and Qwen3.8-27B-FP8 \cite{yang_qwen3_2025}, and Gemma-4-E4B-it and Gemma-4-31B-it-FP8 (RedHatAI/gemma-4-31B-it-FP8-dynamic) \cite{team_gemma_2026, redhatai_gemma4_fp8_2026}. For downstream VLM reasoning experiments (rubric, verifier, and skill-memory evaluation), we use Gemma-4-31B-it-FP8 alone.

\subsection{Probing vision representations}
A linear probe measures how much task-relevant structure is already present in a frozen encoder's latent space, independent of the model's ability to reason about or verbalize it, by fitting a linear classifier to predict classes from a supervised dataset \cite{alain_understanding_2018}. Probing has been used to detect pretraining data \cite{liu_probing_2024} and linguistic properties \cite{conneau_what_2018}. We test whether the vision encoder captures complex features within agricultural images, given that agricultural images were likely a small fraction of these models' training data.

\subsection{Test-time scaling}

Scaling compute at test time, rather than training data or parameter count \cite{snell_scaling_2024}, offers a distinct advantage. Because test-time and pretraining compute are not 1-to-1 ``exchangeable'' (easy-to-medium questions are typically already within a model's capabilities, while harder ones demand more inference-time compute \cite{snell_scaling_2024}), extra compute can be allocated adaptively, exactly where a question needs it. One way to allocate this is to structure reasoning around a fixed, task-specific diagnostic rubric, a chain of criteria mirroring the observations a domain expert would make before reaching a conclusion.

Rather than structuring a single model's own reasoning, test-time compute can instead be spent on model-to-model interaction, where a separate verifier evaluates candidate answers after the fact, though verification effectiveness remains unclear and limited. Typical approaches include LM judges \cite{zheng_judging_2023}, process-supervised reward models \cite{cobbe_training_2021}, and outcome-supervised reward models \cite{lightman_lets_2023}. Kwok et al.\ introduce LLM-as-a-Verifier, a general-purpose, training-free framework that estimates candidate quality via the expectation over scoring-token logits rather than discrete LM-judge scores \cite{kwok_llm-as--verifier_2026}. Across both strategies, the appeal of test-time scaling is recovering performance gain without updating model weights, preserving the base model's capabilities and avoiding the catastrophic forgetting or collapse risk further training can carry.

\subsection{Skill optimization}

A complementary form of test-time scaling accumulates knowledge across conversations rather than within a single response. Instead of reasoning harder on one query, an agent can store what it learns from experience and reuse it on later queries. Tang et al.\ introduce WikiSkill, separating raw execution experience, accumulated knowledge, and executable skills, continuously consolidating experience into a persistent, shared knowledge base that subsequent skill refinements draw on \cite{tang_wikiskill_2026}. We adopt this same principle, in-context accumulation of experience into a persistent, iteratively refined reference artifact, applied to agricultural diagnosis.
\section{Experimental Setup}

\subsection{Evaluation Metrics}
\label{sec:eval_metrics}

\subsubsection{LLM-as-a-judge classification accuracy}
\label{sec:judge_metrics}

Motivated by \cite{ranario_are_2026}, every classification prediction is scored by an LLM judge rather than exact string match, since agricultural diagnosis naming varies across sources and a semantically correct answer rarely matches the ground-truth label's exact wording. Given a prediction $\hat{y}_i$, the ground-truth label $y_i$, and the full list of possible classes for that image's source dataset, the judge returns a match indicator $m_i \in \{0, 1\}$ and a confidence level $\gamma_i \in \{0, 1, 2\}$ (not confident, somewhat confident, very confident). A prediction is counted correct only if the judge both calls it a match and is at least somewhat confident in that call:

\begin{equation}
  c_i = \mathbf{1}[m_i = 1] \cdot \mathbf{1}[\gamma_i \geq \tau],
  \label{eq:judged_correct}
\end{equation}

\noindent with confidence threshold $\tau = 1$. Judge accuracy is the mean of $c_i$ over
the evaluation set. 


\subsubsection{Object detection metrics}
\label{sec:detection_metrics}

Detection predictions are evaluated by greedy IoU-based matching against ground-truth boxes. For predicted boxes and ground-truth boxes in an image, each candidate pair is scored by standard axis-aligned intersection-over-union, and pairs are matched in descending IoU order, each box used at most once, keeping only pairs with $\mathrm{IoU} \geq 0.5$. Unmatched predictions are false positives and unmatched ground-truth boxes are misses. We report standard precision, recall, and F1 computed from the resulting true positive, false positive, and false negative counts, macro-averaged per image.

\subsection{Linear probing setup}
\label{sec:linear_probe}

Let $\mathcal{D} = \{(x_i, y_i)\}_{i=1}^{N}$ be a dataset of agricultural images $x_i$ with class labels $y_i \in \{1, \ldots, K\}$ with $K$ the pooled number of classes across all datasets. We freeze all parameters of the vision encoder $V$ and register a forward hook at its final layer, before the vision-language projector, to capture patch-level feature maps. For each image, the hook yields a sequence of $P$ patch tokens where $d$  is the dimension size for each patch:

\begin{equation}
  \mathbf{H}_i = V(x_i) \in \mathbb{R}^{P \times d}
\end{equation}

\noindent We apply mean pooling over the patch sequence to obtain a single fixed-size image embedding:

\begin{equation}
  \mathbf{v}_i = \frac{1}{P} \sum_{p=1}^{P} \mathbf{H}_i^{(p)} \in \mathbb{R}^{d}
\end{equation}

\noindent A single fully-connected linear layer $F_\theta : \mathbb{R}^{d} \rightarrow \mathbb{R}^K$ is trained on the frozen embeddings by minimizing the cross-entropy loss:

\begin{equation}
  \min_{\theta} \; \sum_{i=1}^{N}
  \mathcal{L}_{\mathrm{CE}}\!\left(F_\theta(\mathbf{v}_i),\; y_i\right)
\end{equation}

\noindent $F_\theta$ is optimized with Adam ($\mathrm{lr} = 10^{-3}$, batch size $= 256$) for 50 epochs with all backbone parameters frozen. We report separability as macro-averaged F1: per-class F1 computed from the predictions $\hat{y}_i = \arg\max_c F_\theta(\mathbf{v}_i)_c$ against $y_i$, averaged uniformly over the $K$ classes.

Fitting a linear probe requires a supervised dataset (input and labeled output). We used labeled data from each dataset that is outside the sampled test set. We deliberately restrict the probe to a single linear layer with no hidden units ($d \times K + K =$ num. of parameters) to ensure that classification performance reflects the quality of the frozen visual representations rather than the capacity of the classifier. A high-capacity head could compensate for poor features, causing the probing metrics to be unfaithful to visual separability.

We include these purely vision-trained models, DINOv3 and YOLOv11, to test whether their feature spaces exhibit better class separability than the VLM's own vision encoder, which is trained under a contrastive, language-aligned objective rather than for discriminative visual recognition. This comparison lets us isolate whether the VLM's vision encoder itself is a limiting factor in downstream task performance, or whether the bottleneck lies elsewhere in the pipeline.

\begin{figure}[t]
    \centering
    \includegraphics[width=1.0\linewidth]{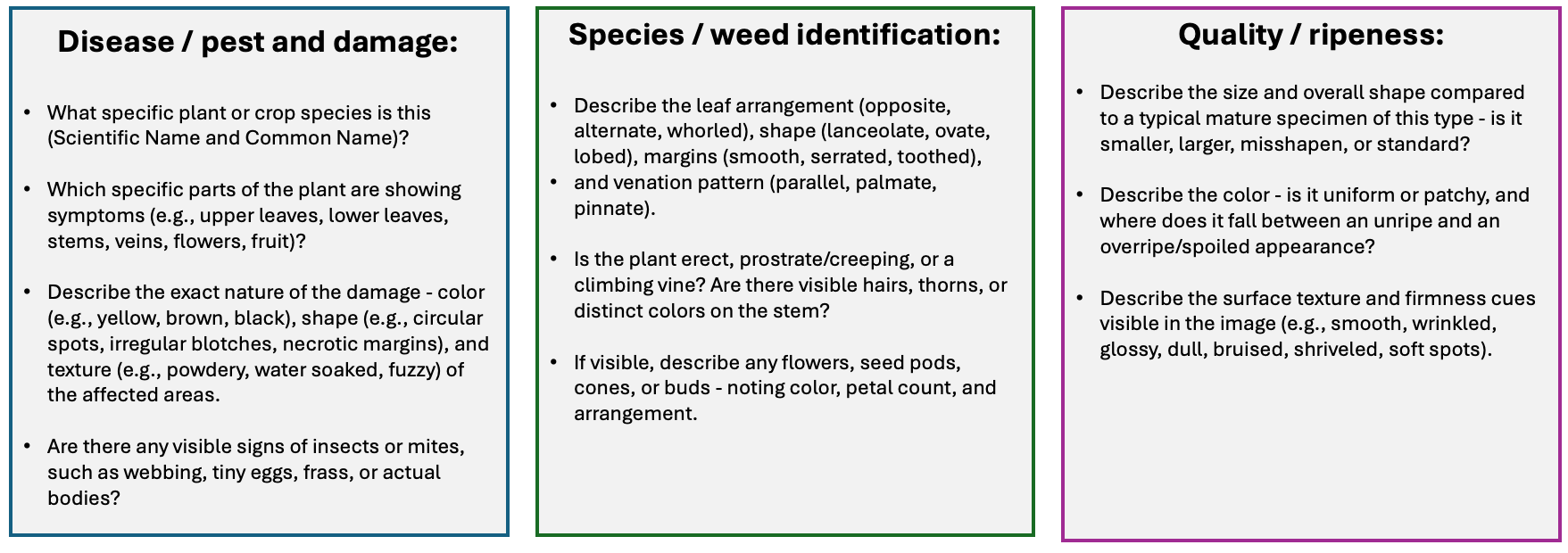}
    \caption{Fixed diagnostic criteria chains used as the shared rubric for structured candidate generation and pairwise verification, for each task type. The model answers each criterion in sequence, appended to the base classification question.}
    \label{fig:cot_context}
\end{figure}

\begin{figure}[t]
    \centering
    \includegraphics[width=1.0\linewidth]{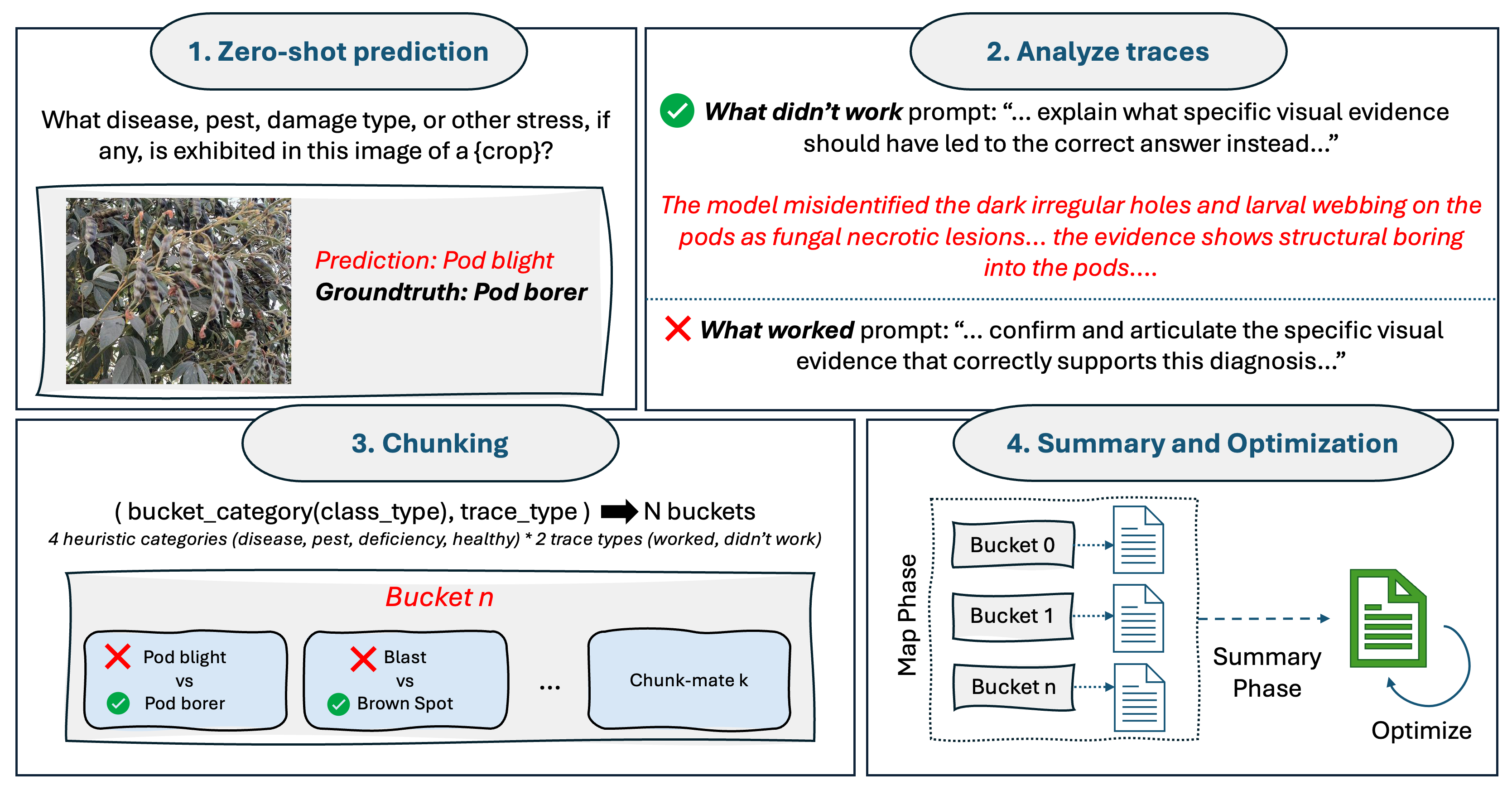}
    \caption{The skill-memory optimization pipeline. (1) Zero-shot prediction on the \emph{lessons} set produces a trace per image. (2) Each trace is explained via a "what worked" or "what didn't work" prompt depending on correctness. (3) Explanations are bucketed by (category, trace type). (4) Each bucket is summarized (Map Phase), merged into a single skill document (Summary Phase), and then iteratively edited against resampled correct/incorrect examples over $R$ rounds (Optimize).}
    \label{fig:flowchart}
\end{figure}

\subsection{Rubric and verifier setup}

Building on the LLM-as-a-Verifier framework described above \cite{kwok_llm-as--verifier_2026}, we structure test-time verification around a fixed, task-specific diagnostic rubric, applied to three task types: disease/pest and damage, species/weed identification, and quality/ripeness. For each task, the rubric is a manually authored chain of criteria mirroring the sequence of observations a domain expert would make before reaching a diagnosis (Figure~\ref{fig:cot_context}).

For each image, the model generates $k$ candidate responses by answering the rubric in a single fill-in-the-blank turn (rather than across separate conversational turns, for efficient $K$-sampling), each sampled independently at temperature $T=0.8$. The pool is ranked by one pass of the Probabilistic Pivot Tournament (PPT) \cite{kwok_llm-as--verifier_2026}, and the top-ranked candidate is reported as the final prediction; see that paper for the full derivation.

We additionally evaluate the same rubric as a standalone single-pass prompting condition (Manual rubric CoT), isolating what the rubric alone contributes independent of candidate generation and verification.

\subsection{Skill optimization loop}
\label{sec:skill_optimization}

As a counterpart to the verifier framework above, we test our second hypothesis directly: whether accumulating diagnostic knowledge in context improves agricultural classification performance. For each task $t$, let $\mathcal{D}_t$ be a held-out pool of images disjoint from both the model's standard evaluation split and from the other tasks' pools: an external Digital Green corpus \cite{digigreen_crop_disease_2026} for disease/pest and damage, five held-out AgML quality-grading datasets for quality/ripeness, and iNatAg-mini \cite{jain_inatag_2025} for species/weed identification. We partition $\mathcal{D}_t$ 80/20 into a lessons set $\mathcal{D}_t^{\mathrm{lessons}}$, used to build and edit a skill-memory document, and a validation set $\mathcal{D}_t^{\mathrm{val}}$, used only to score it. The same model $M$ is used for generation, judging, and editing. A summary of the optimization loop can be seen in Figure \ref{fig:flowchart}.

\textbf{1. Zero-shot prediction:} For each $x_i \in \mathcal{D}_t^{\mathrm{lessons}}$, the model produces a prediction $\hat{y}_i = M(x_i)$ to the base classification question (e.g. ``What disease, pest, damage type, or other stress, if any, is exhibited in this image of a \{crop\}?''), judged against the ground truth via the LLM judge of Section~\ref{sec:judge_metrics}, producing $c_i$ (Equation~\ref{eq:judged_correct}). 

\textbf{2. Analyze traces:} Each $(x_i, \hat{y}_i, y_i, c_i)$ is passed to one of two prompts that reveal $y_i$ and elicit an explanation $e_i$. When $c_i = 0$, the \textit{what didn't work} prompt asks the model to explain what visual evidence should have led to $y_i$ instead of $\hat{y}_i$; when $c_i = 1$, the \textit{what worked} prompt asks it to confirm and articulate the visual evidence supporting $y_i$, explicitly distinguishing sound reasoning from a right answer reached for the wrong reason.

\textbf{3. Chunking:} Each explanation $e_i$ is assigned to a bucket $b_i = (\mathrm{cat}(y_i),\, c_i)$, where $\mathrm{cat} : \mathcal{Y} \rightarrow \mathcal{C}$ is a task-specific heuristic grouping function. For disease/pest and damage, $\mathcal{C} = \{\text{disease}, \text{pest}, \text{deficiency}, \text{healthy}\}$, giving $|\mathcal{C}| \times 2 = 8$ buckets; for species, $\mathcal{C} = \{\text{plant}, \text{weed}\}$ gives 4 buckets; for quality, $\mathcal{C}$ is a single category, since the five source datasets' grading vocabularies (e.g. grade-1/A/ripe) share no keyword-detectable taxonomy to split on further, giving 2 buckets. Although disease and pest/damage share a single lessons pool and optimization run, Table \ref{tab:method_by_task} and Table \ref{tab:seen_unseen} report judge accuracy for each separately on the main benchmark evaluation set.

\textbf{4. Summary and optimization:} In a Map Phase, each bucket's explanations $\{e_i : b_i = b\}$ are summarized by $M$ into a short list of recurring, generalizable visual patterns, not memorized per-image details. A Summary Phase merges these per-bucket summaries into a single skill-memory document $S^{(0)}$, with an empty initial changelog $\mathrm{changelog}^{(0)}$. $S^{(0)}$ is then iteratively edited over $R = 10$ rounds. In round $r$, $S^{(r)}$ is applied as a system-prompt addition to re-score $\mathcal{D}_t^{\mathrm{lessons}}$. Up to $n = 10$ correct and $n = 10$ incorrect examples per bucket are resampled from this re-scored pool, denoted $\{e_i\}_{\mathrm{resampled}}^{(r)}$. And an edit call, shown both this resampled set and $\mathrm{changelog}^{(r)}$ (the natural-language record of what changed and why when $S^{(r)}$ was itself produced), revises the document and produces a new changelog documenting this round's own edit:

\begin{multline}
  \big(S^{(r+1)},\; \mathrm{changelog}^{(r+1)}\big) = \mathrm{Edit}\big(S^{(r)}, \\
  \{e_i\}_{\mathrm{resampled}}^{(r)},\; \mathrm{changelog}^{(r)}\big).
\end{multline}

\noindent Each round's document is scored by judge accuracy using a judge $J$, computed over $\mathcal{D}_t^{\mathrm{val}}$ with $S^{(r)}$ applied as the system-prompt addition, written $J(S^{(r)})$, and the best-scoring round

\begin{equation}
  S^{\star} = \arg\max_{r \in \{0,\ldots,R\}} J\big(S^{(r)}\big)
\end{equation}

\noindent is kept and injected as a system-prompt addition for every subsequent inference call on task $t$, requiring no gradient updates to the model $M$ itself.

\begin{figure}[t]
    \centering
    \includegraphics[width=0.5\linewidth]{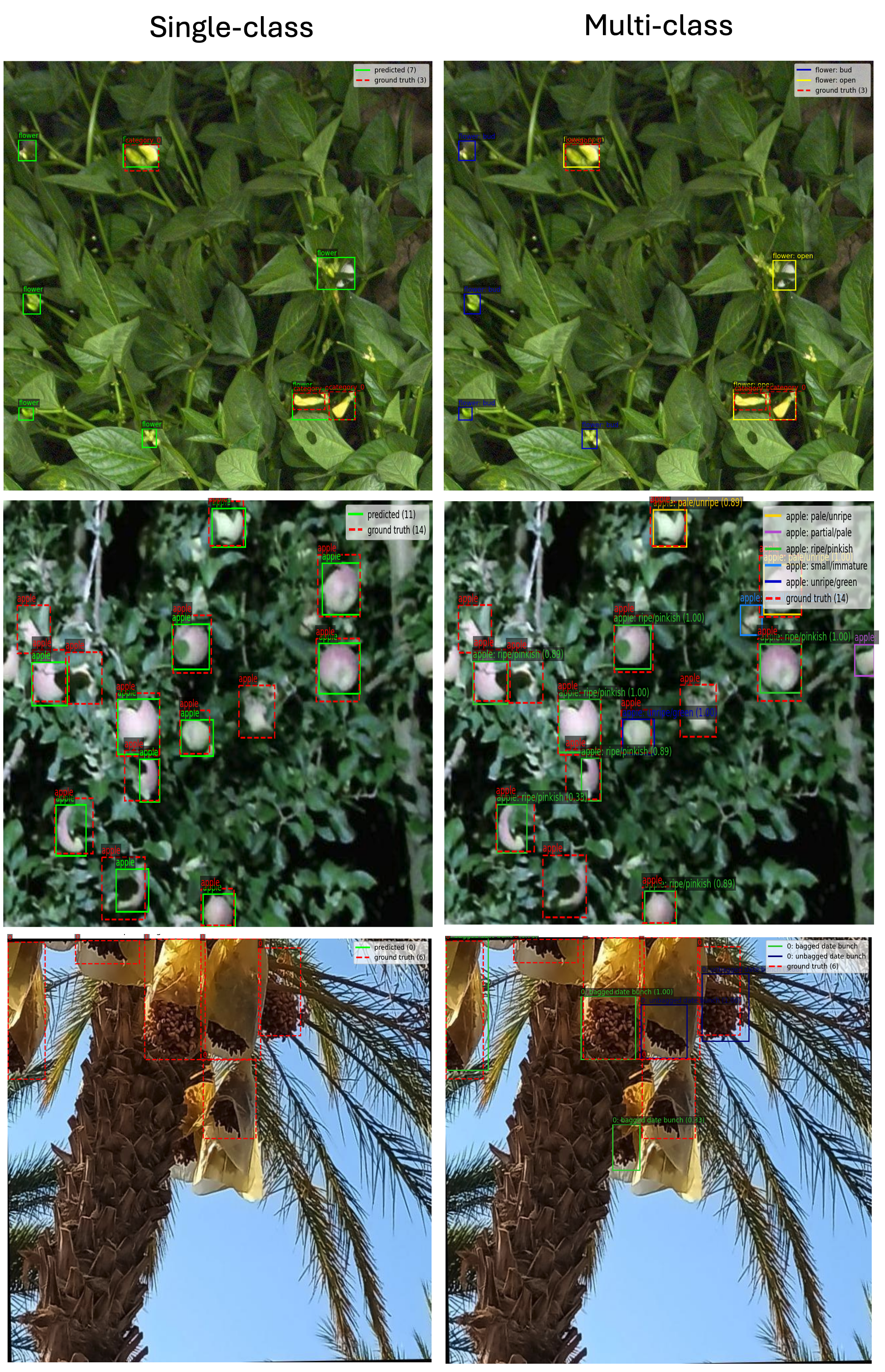}
    \caption{Qualitative comparison of single-class (left) vs. multi-class (right) detection on three example images, showing predicted boxes and labels under the single-class and multi-class conditions.}
    \label{fig:multi_class}
\end{figure}

\subsection{Object detection setup}
\label{sec:detection_setup}

We evaluate object detection under five conditions. Baseline detection prompts the model with one category at a time, localizing every visible instance of that single class per pass. \textit{Multi-class detection} instead asks for every class in the dataset in a single pass, with each instance labeled by both its category and a free-text sub-type (e.g. growth stage, ripeness) the model observes directly. A predicted box is matched to a ground-truth box at $\mathrm{IoU} \geq 0.5$. Among matched pairs, we further distinguish a tight match ($\mathrm{IoU} \geq 0.7$) from a loose match ($0.5 \leq \mathrm{IoU} < 0.7$), and separately track unmatched predictions (false positives) and unmatched ground-truth instances (misses). Box matching uses IoU alone.

\textbf{Multi-class + filtering} selects, per image, whichever single predicted sub-class category yields the highest F1 against that image's own ground truth, scoring only that sub-class's boxes. This is an oracle upper bound. The selection itself consults ground truth, which no deployable filter could do, so it reports the best achievable result if sub-class selection were perfect rather than a method's own performance.

\textbf{Multi-class + skill} extends the skill-optimization loop of Section~\ref{sec:skill_optimization} to detection. The held-out pool is drawn from 15 datasets disjoint from the 54-dataset detection benchmark used elsewhere in this paper (10 for the lessons set, 5 for validation), split by whole dataset rather than by image, so no image from a lessons-set dataset ever appears in validation or in Table 3's benchmark. Traces are analyzed per box rather than per image. Each false positive, miss, loose match, and tight match (``worked'') is explained individually, then chunked directly by this four-way error type (no further class-type split, since error type alone already encodes the failure/success mode). The Map and Summary Phases proceed as in Section~\ref{sec:skill_optimization}, and the same ten-round edit loop applies, with the reward $J(S^{(r)})$ computed as F1 at $\mathrm{IoU} \geq 0.5$ on the held-out validation datasets in place of accuracy. No LLM judge is needed for detection, since box overlap is scored directly.

\textbf{Multi-class + skill + filtering} combines both. The best-scoring skill document from the optimization loop is injected as a system-prompt addition, and oracle sub-class filtering is applied on top of its predictions.
\section{Results and discussion}

\begin{figure}[t]
    \centering
    \includegraphics[width=0.9\linewidth]{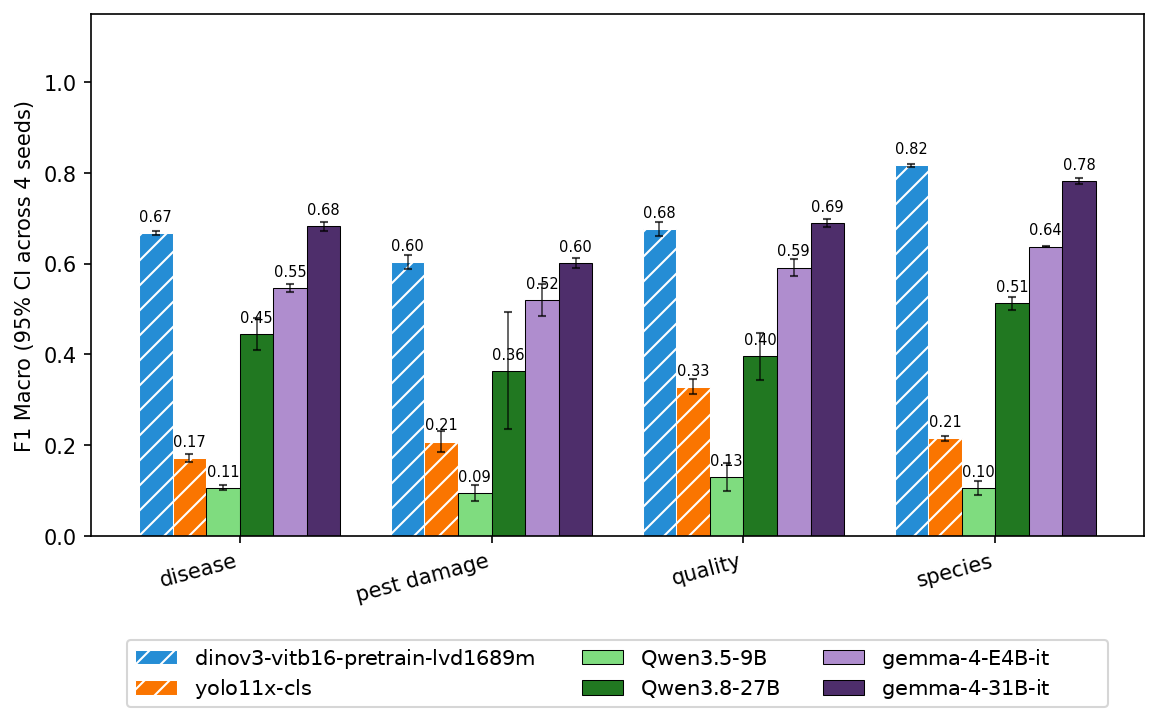}
    \caption{Linear probe F1-macro separability by backbone and task category, averaged over 4 seeds with 95\% confidence intervals.}
    \label{fig:probe_summary}
\end{figure}

\subsection{Vision towers encode sufficient information}

Figure \ref{fig:probe_summary} reports F1-macro separability for each backbone across the four task categories, averaged over 4 seeds with 95\% confidence intervals. DINOv3 achieves high separability on every task (disease: 0.67, pest/damage: 0.60, quality: 0.68, species: 0.82), consistent with its self-supervised pretraining on a larger, more diverse corpus than YOLO11x-cls's training corpus, which trails by a wide margin on every task (0.17--0.33). Among the VLM vision encoders, Qwen3.5-9B underperforms (0.09--0.13), while Qwen3.8-27B-FP8 recovers substantially (0.36--0.51). Gemma 4's encoders are the strongest VLM backbones tested, with Gemma-4-31B-it-FP8 approaching or matching DINOv3 on every task (disease: 0.68, pest/damage: 0.60, quality: 0.69, species: 0.78) and Gemma 4 E4B-it close behind (0.55--0.64). This indicates the vision encoder itself is not the primary bottleneck. Any degradation in end-to-end classification is more plausibly attributable to how well these visual features are aligned with and surfaced through the language modeling pathway than to a representational limitation of the vision encoder. The language components built on top of it may lack the domain-specific terminology and taxonomic knowledge to name what they are seeing correctly. This reframes the problem of improving agricultural VLM performance as one of supplying the model with the right vocabulary and knowledge to map those features onto the correct diagnosis. In other words, these models see substantially more about agricultural conditions than they ultimately report, the first half of the claim motivating this work.  Note that F1-macro from a closed-set, supervised linear probe and open-ended, zero-shot judge accuracy are not directly comparable metrics. The gap between them is suggestive evidence for a downstream alignment bottleneck, not a precise measurement of it.

\begin{figure}[t]
    \centering
    \includegraphics[width=1.0\linewidth]{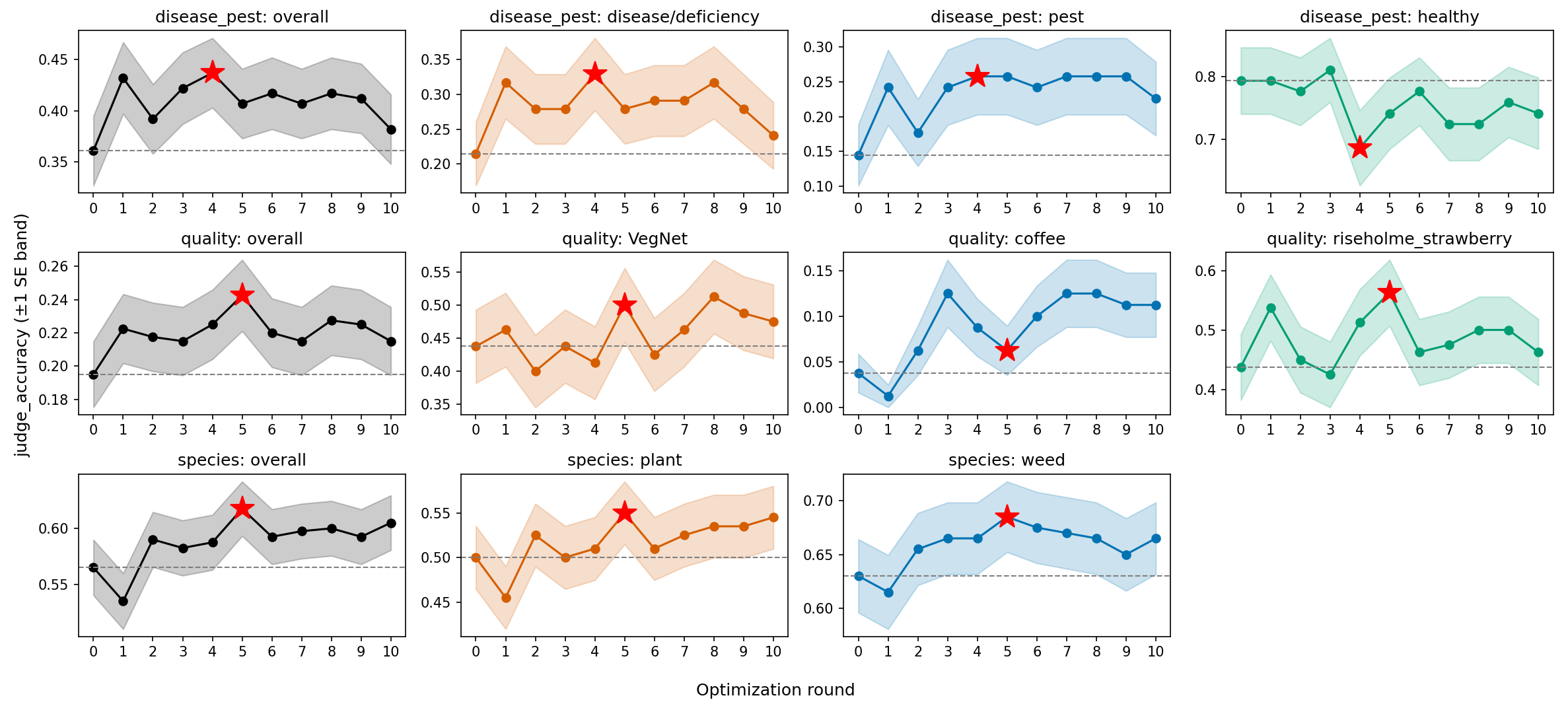}
    \caption{Judge accuracy across the $R=10$ skill-memory optimization rounds, shown per task and per bucket (overall and class-level subsets), averaged with $\pm 1$ SE bands over held-out validation data. The red star marks the best-scoring round $S^\star$ selected; the dashed horizontal line marks round-0 (pre-optimization) performance.}
    \label{fig:train_graphs}
\end{figure}

\begin{table}[t]
\centering
\footnotesize
\setlength{\tabcolsep}{4pt}
\begin{tabular}{lcccc}
\toprule
\textbf{Method} & \textbf{Disease} & \textbf{Pest} & \textbf{Quality} & \textbf{Species}
\\
\midrule
Baseline & 0.297 & 0.185 & 0.403 & 0.195 \\
Rubric CoT & 0.230 & 0.111 & 0.359 & 0.189 \\
PPT verifier & 0.295 & 0.173 & 0.407 & \textbf{0.199} \\
With skill & \textbf{0.302} & \textbf{0.216} & \textbf{0.414} & 0.172 \\
\bottomrule
\end{tabular}
\caption{AgML classification accuracy by method and task type (Gemma-4-31B-it-FP8). CoT = manual rubric chain-of-thought; PPT verifier = LLM-as-a-verifier with the Probabilistic Pivot Tournament. Best result per task in \textbf{bold}.}
\label{tab:method_by_task}
\end{table}

\subsection{Test-time scaling and the limits of parametric knowledge}

Across both classification and detection, test-time scaling gains over baseline are modest and inconsistent (Tables~\ref{tab:method_by_task}, \ref{tab:detection_comparison}). Classification's PPT-verifier condition shows small, sometimes negative, shifts over baseline, but Rubric CoT alone drops substantially on disease and pest (6.7 and 7.4 points below baseline, respectively), indicating the rubric's fixed structure can actively hurt single-pass prediction even where the verifier recovers most of that loss. Among non-oracle detection conditions, injecting skill-memory alone (without multi-class prompting or filtering) is the best, improving F1 by only 0.009 over baseline (0.421 vs.\ 0.412). Multi-class prompting alone does not help (0.400 vs.\ 0.412), and adding skill-memory to it recovers only part of that gap (0.402). Notably, multi-class + filtering, the oracle upper bound, reaches the highest F1 of any condition (0.432), well above what any instruction-only method achieves on its own. As expected, filtering raises precision substantially (0.408 to 0.541, and 0.413 to 0.548 with skill-memory added) at some cost to recall, since selecting against ground truth suppresses false positives rather than recovering missed detections. This gap indicates the model is already capable of producing a correct sub-class label for a given box, but the unconstrained prompt does not reliably surface it, pointing to the instruction and selection mechanism, not model capacity, as the binding constraint.

Since none of these methods update model weights, they can only restructure or verify knowledge the base model already holds parametrically, not manufacture new knowledge. This bound is visible directly in Table~\ref{tab:seen_unseen} for classification tasks. Disease classes seen during skill-optimization, even on a held-out dataset, improve at a far higher rate than unseen classes (42.3\% vs.\ 22.4\%), and unseen classes show a net regression (45 worse vs.\ 43 improved), suggesting the gain reflects refined use of existing exposure rather than genuinely new knowledge. The same limitation extends to detection's free-text sub-type labels, generated with no external grounding. Pairing either task with a standardized taxonomic knowledge base, or for detection, interaction with open-vocabulary models supplying grounded candidate labels, would shift both from unconstrained generation toward retrieval-and-verification, a direction current parametric-knowledge-bound methods cannot reach on their own.

\begin{table}[t]
\centering
\footnotesize
\begin{tabular}{llcccc}
\hline
\textbf{Task type} & \textbf{Subset} & \textbf{$n$ classes} & \textbf{Improved} &
\textbf{Worse} \\
\hline
disease      & seen   & 26  & \textbf{\color{ForestGreen}{11}}  & 8 \\
disease      & unseen & 192 & 43 & \textbf{\color{red}{45}} \\
pest\_damage & seen   & 6   & \textbf{\color{ForestGreen}{3}} & 1 \\
pest\_damage & unseen & 78  & \textbf{\color{ForestGreen}{17}} & 10 \\
\hline
\end{tabular}
\caption{Per-class improvement after skill-memory injection (Gemma-4-31B-it-FP8) for classification, split by whether the class was seen during optimization (in the \emph{lessons} set) or unseen. Seen classes improve at a far higher rate than unseen classes, which show a net regression, indicating the gain reflects refined use of existing exposure rather than new knowledge. Class count reflects subset of classes appearing in the held-out skill-memory evaluation set.}
\label{tab:seen_unseen}
\end{table}


\subsection{Toward agentic knowledge access}

Recent agentic frameworks extend what a model can do at inference time by giving it a harness, tools, and access to external resources, rather than relying solely on what is stored in its weights. This suggests a direction beyond the in-context skill-memory we evaluate here. Rather than compressing lessons learned into a single static document injected at inference time, a model could be given tool access to a standardized, external knowledge base, taxonomic references for species identification, structured disease/pest ontologies, or similar, and use it as a reliable source to consult rather than a fixed document to memorize from. This reframes the model's role from answering from memory to problem-solving with reliable information actively retrieved, which our static skill-memory results suggest is necessary given the limits described above.



\begin{table}[t]
\centering
\footnotesize
\setlength{\tabcolsep}{4pt}
\begin{tabular}{lccc}
\hline
\textbf{Condition} & \textbf{Precision} & \textbf{Recall} & \textbf{F1} \\
\hline
Baseline detection & 0.426 & 0.466 & 0.412 \\
Baseline + skill & 0.446 & \textbf{0.473} & 0.421 \\
Multi-class detection & 0.408 & 0.472 & 0.400 \\
Multi-class + filtering & 0.541 & 0.424 & \textbf{0.432} \\
Multi-class + skill & 0.413 & 0.468 & 0.402 \\
Multi-class + skill + filtering & \textbf{0.548} & 0.415 & 0.427 \\
\hline
\end{tabular}
\caption{Object detection performance by condition (Gemma-4-31B-it-FP8). Multi-class + filtering is an oracle upper bound using ground-truth-informed sub-class selection; Multi-class + skill and Multi-class + skill + filtering use the best-scoring skill document from the optimization loop.}
\label{tab:detection_comparison}
\end{table}

\section{Limitations and conclusion}

Our LLM-judge evaluation needs further validation. Its scoring criteria trade off generality against strictness (a lenient judge inflates accuracy on loosely or ambiguously named categories, e.g. disease and pest terminology that varies by region, while a stricter judge penalizes correct but differently phrased predictions), and we did not systematically quantify this trade-off or calibrate it against human annotations. Additionally, the same model serves as both the system under evaluation and the judge throughout skill optimization, so judge accuracy may be inflated by shared failure modes or self-consistency bias rather than reflecting true correctness against an independent standard.

The skill-memory optimization procedure could also have been refined further. The number of examples resampled per bucket at each edit round was capped at 10 by default and never ablated, by default and never ablated, and we observed the model sometimes adheres too strictly to the skill document's own vocabulary, favoring a term it introduced over the evaluation dataset's actual label scheme, turning a confidently stated skill into a source of error on datasets it was never built from. More importantly, Figure~\ref{fig:train_graphs} shows optimization can overfit to the held-out validation set used for round selection, with later rounds improving validation accuracy without reliably transferring to test, indicating our round-selection criterion is itself vulnerable to the same generalization gap skill-memory is meant to address. Taken together, our first hypothesis holds, vision is good, while our second does not. Accumulating knowledge in context is not consistently good, leaving test-time scaling bounded by what the model already parametrically knows rather than a reliable path to new knowledge.
{
    \small
    \bibliographystyle{ieeenat_fullname}
    \bibliography{main}
}

\end{document}